\documentclass{article}

\usepackage{arxiv}

\usepackage[utf8]{inputenc} 
\usepackage[T1]{fontenc}    
\usepackage{hyperref}       
\usepackage{url}            
\usepackage{booktabs}       
\usepackage{amsfonts}       
\usepackage{nicefrac}       
\usepackage{microtype}      
\usepackage{lipsum}		
\usepackage{graphicx}
\usepackage[square,sort,comma,numbers]{natbib}
\usepackage{doi}
\usepackage{amsmath}

\title{Dual-Path Region-Guided Attention Network for Ground Reaction Force and Moment Regression}

\author{Xuan Li
\\
	Department of Biomedical Engineering\\
	Johns Hopkins University\\
	Maryland, MD 21210 \\
	\texttt{xli375@jh.edu} \\
	\And
    Samuel Bello \\
	Department of Biomedical Engineering\\
	Johns Hopkins University\\
	Maryland, MD 21210 \\
	\texttt{sbello2@jhu.edu} \\
}

\renewcommand{\shorttitle}{\textit{arXiv} Template}

\hypersetup{
pdftitle={A template for the arxiv style},
pdfsubject={q-bio.NC, q-bio.QM},
pdfauthor={David S.~Hippocampus, Elias D.~Striatum},
pdfkeywords={First keyword, Second keyword, More},
}

\begin{document}
\maketitle

\begin{abstract}
Accurate estimation of three-dimensional ground reaction forces and moments (GRFs/GRMs) is crucial for both biomechanics research and clinical rehabilitation evaluation.
In this study, we focus on insole-based GRF/GRM estimation and further validate our approach on a public walking dataset. We propose a Dual-Path Region-Guided Attention Network that integrates anatomy-inspired spatial priors and temporal priors into a region-level attention mechanism, while a complementary path captures context from the full sensor field. The two paths are trained jointly and their outputs are combined to produce the final GRF/GRM predictions. Conclusions: Our model outperforms strong baseline models, including CNN and CNN–LSTM architectures on two datasets, achieving the lowest six-component average NRMSE of 5.78 \% on the insole dataset and 1.42 \% for the vertical ground reaction force on the public dataset. This demonstrates robust performance for ground reaction force and moment estimation.
\end{abstract}

\keywords{Ground Reaction Force \and Attention Mechanism \and Spatiotemporal modeling \and Gait Analysis}

\section{Introduction}
The Ground Reaction Force (GRF) and Ground Reaction Moment (GRM) are indispensable metrics for kinetic analysis. GRF is the force exerted by the ground on the foot during contact with it, and GRM represents the torque generated by these forces \citep{winter2009biomechanics}. By quantifying the forces and moments, we can derive joint kinetics through inverse dynamics \citep{ozeloglu2024combining}, \citep{SUTHERLAND2005447}. In recent years, researchers have been investigating the clinical assessment of neurological diseases such as stroke and Parkinson's disease, as well as for analyzing diverse movement patterns \citep{xiang2024rethinking} \citep{caramia2018imu}. Additionally, it has been used for the rehabilitation of musculoskeletal injuries, including Anterior Cruciate Ligament (ACL) tears \citep{jafarnezhadgero2021analysis}. Furthermore, in sports science, analyzing GRF helps identify the underlying mechanisms to enhance athletic outcomes, such as detecting the movement patterns that may lead to injuries, or as a guide to improve performance \citep{bezodis2017alterations}. Consequently, GRF/GRM data are vital for both clinical decision-making and the optimization of performance in sports science.

The traditional way to collect GRF/GRM data uses specialized force plates. Although this gold standard provides accurate measurements, it is mainly limited to lab environments. A critical limitation of the force plate is the lack of ambulatory ability. Force plates remain static measurement tools and cannot continuously evaluate an individual walking freely during daily, outdoor activities. This constraint makes it difficult to collect real-world movement data. Many studies have explored wearable kinetic measurement systems as alternatives to address this issue. For example,  \citep{day2021low} demonstrated that IMU-derived estimates of peak vertical GRF and contact time are sensitive to signal filtering parameters, highlighting the challenge of achieving accurate force estimation from wearable data; \citep{sakamoto2021validity} developed a thin and flexible stretch strain sensor that allows foot motion analysis during walking and running without the need for cameras. There are many recent works using sensorized insole to evaluate gait \citep{nascimento2022development} \citep{pergolini2024assessment} \citep{crea2014wireless} \citep{agarwal2020wireless} \citep{ivanov2019design}.  However, these approaches often suffer from low sensor density, which can fail to capture detailed plantar pressure distributions, leaving a gap in accurately estimating kinetic signals from wearable data. To directly address the data quality bottleneck while maintaining real-world usability, our work utilizes a novel high-density insole pressure sensor system. This system generates time-series data that is inherently spatially-dependent and non-Euclidean, a format that requires specialized processing beyond standard methods.

While these flexible and portable sensors increase access to gait measurement, accurately interpreting the collected signals remains challenging. Deep learning provides a powerful framework to model these high-dimensional and nonlinear dynamics, enabling end-to-end mapping from wearable sensor data to biomechanical outcomes. However, most existing works has been using IMU-based sensors or low-density insoles. CNN-based models have been popular for GRFs and GRMs regression \citep{johnson2018predicting} \citep{chen2025prediction} \citep{dorschky2020cnn}. These models are first implemented in images, designed for regular grid structures, which limits their ability to capture complex spatial dependencies among sensor elements that show non-uniform distributions. Furthermore, few methods explicitly leverage anatomical region priors combined with sensor spatial topology information, treating the complex interaction between feet and the ground as an unstructured image. To overcome the challenges of non-uniform sensor layouts, along with the small sample sizes of laboratory data, and the high cost of large-scale data collection, it's crucial to leverage prior knowledge to bridge this gap. We aim to develop a model that minimizes reliance on data volume by emphasizing structural constraints, allowing it to achieve a robust balance between data-driven learning and knowledge-based prior constraints.

In this paper, we propose a Dual-Path Region-Guided Attention Network (DP-RGNet), in which one path leverages anatomy-informed plantar regions to organize local feature aggregation, while the other path processes globally pooled representations without explicit regional constraints. This dual-path design allows the model to balance structured region guidance with unconstrained context, leading to significantly improved GRF/M prediction accuracy. The main contributions are:

\paragraph{1.}We propose a novel Dual-Path Region-Guided Attention Network (DP-RGNet) architecture specifically designed for mapping high-dimentional plantar pressure data to GRF/GRM kinetics that requires laboratory instrument to measure.

\paragraph{2.}The DP-RGNet incorporates both a spatial positional encoding (based on coordinates) and a dynamic positional encoding (based on the Center of Pressure) combined with a region mask attention mechanism, significantly outperforming common baseline models like CNN+LSTM across different datasets.

\paragraph{3.}The proposed model is evaluated on both a custom wearable insole dataset and a public walking dataset collected using a pressure sensing mat. The validation across different measurement systems demonstrates the model's generalizable and stable performance.

\section{Experimental Methods}

\subsection{Dataset Overview}
The study is based on two datasets. A dataset using self-collected high-density instrumented insole system to collect input data, which consists of $64 \times 16$ individual sensing elements, capturing plantar pressure distribution at sampling frequency of $40 \text{ Hz}$. And a public dataset featuring high-resolution plantar pressure measurements with footsteps in size of $75 \times 40$ at sampling frequency of $101 \text{ Hz}$.

\subsection{Dataset A: Custom insole dataset}

\subsubsection{Data Collection Protocol}
Data was collected from four healthy subjects (Details see \textbf{Table 1}) walking on a force-plate-instrumented treadmill. The force plate provided ground truth 6-axis GRF/GRM data, synchronized with the insole pressure data (sampled at $40$ $\text{Hz}$ and $100$ $\text{Hz}$ respectively). The gait protocol was a walking-only matrix across a range of speeds: $0.75 \text{ m/s}$ and $1.0 \text{ m/s}$ (All 4 subjects), $1.5 \text{ m/s}$ (Subjects 1, 2, 3), $2.0 \text{ m/s}$ (Subjects 2, 3). Subject 1 performed $120 \text{ seconds}$ of walking per trial, while Subjects 2, 3, and 4 performed $240 \text{ seconds}$ per trial, resulting in 2,343 total annotated steps.

\begin{table}[h]
\centering
\caption{Anthropometric details for the dataset A}
\label{tab:subject_details}
\begin{tabular}{c c c c}
\toprule
\textbf{Subject ID} & \textbf{Height (mm)} & \textbf{Weight (kg)} & \textbf{Age (years)} \\
\midrule
1 & 1752.6 & 77.11 & 73 \\
2 & 1828.8 & 78.47 & 64 \\
3 & 1778.0 & 71.70 & 26 \\
4 & 1600.2 & 70.30 & 32 \\
\bottomrule
\end{tabular}
\end{table}

\subsubsection{Preprocessing Pipeline}
\paragraph{Feature Extraction}
The plantar pressure signals acquired from the insole sensor were preprocessed by segmenting individual strides. A stride was defined as the interval between two consecutive heel strikes of the same foot. Heel strikes were detected using a threshold set at 12.5\% of the range of the mean pressure across all sensors. For each peak–trough pair within a trial, points exceeding the threshold with a positive slope were labeled as heel strikes, while those crossing it with a negative slope corresponded to toe-offs. The identified toe-off events delineated the stance and swing phases of the gait cycle. Only the stance-phase data, when the foot maintained ground contact, were retained for kinematic estimation.
\paragraph{Temporal Synchronization and Filtering}
The insole and force plate data streams were temporally synchronized by aligning the heel strike events detected by each system. This ensures the accurate correspondence between the ground truth GRF/GRM and the insole pressure signals. The ground truth signals were then smoothed using a second-order low-pass Butterworth filter with a cutoff frequency of $10 \text{ Hz}$ to remove high-frequency noise while preserving biomechanically relevant kinetics. 
\paragraph{Normalization}
To mitigate the influence of inter-subject variability, GRFs are normalized by the subjects' body weight, and GRMs are normalized by the product of body weight and height. This converts the raw measurements into uniform units, ensuring fair comparison across subjects with different body types.
To account for step duration variability, each extracted stance phase was temporally normalized to a fixed length of $L=40$ time points using cubic spline interpolation. This provides a uniform input dimension for the subsequent DP-RGNet model.

\subsection{Dataset B: UNB StepUP-P150}
UNB StepUP-P150 dataset\citep{larracy2025dataset} collected 150 subjects. The data was acquired using a large $1.2~m \times 3.6~m$ pressure-sensing walkway, providing an exceptional spatial resolution of $5 \times 5$ mm. For this study, we focused on a specific subset of the collection protocol. We included data from all 150 participants, restricting to the barefoot (BF) footwear condition. The subset covers all four walking speed tasks tested in the original protocol: Preferred Speed (W1), Slow-to-Stop (W2), Slow Speed (W3), and Fast Speed (W4). This selection was made to mirror the physical interface of an insole, where the foot is in direct contact with the sensor, allowing for a valid comparison of model performance between the two different measurement systems capturing the same movement phenomenon. 

For applications in machine learning and deep learning, the dataset provides preprocessing pipelines to extract footstep segments. The pipeline standardized footstep representations while retaining crucial information related to foot size and body weight. The footstep was translated to align the centroid of its bounding box to a consistent spatial dimension of $75 \times 40$ pixels.
\paragraph{Spatial Translation} The footstep was rolled to the zero to align the centroid of its bounding box with the center of the padded area.
\paragraph{Temporal Interpolation} To standardize the duration of the gait cycle, nearest-neighbor interpolation was applied to normalize all footsteps to a consistent length of 101 frames. Each frame subsequently represents an approximately $1\%$ interval of the whole stance phase.

\begin{figure}
    \centering
    \includegraphics[width=1\textwidth]{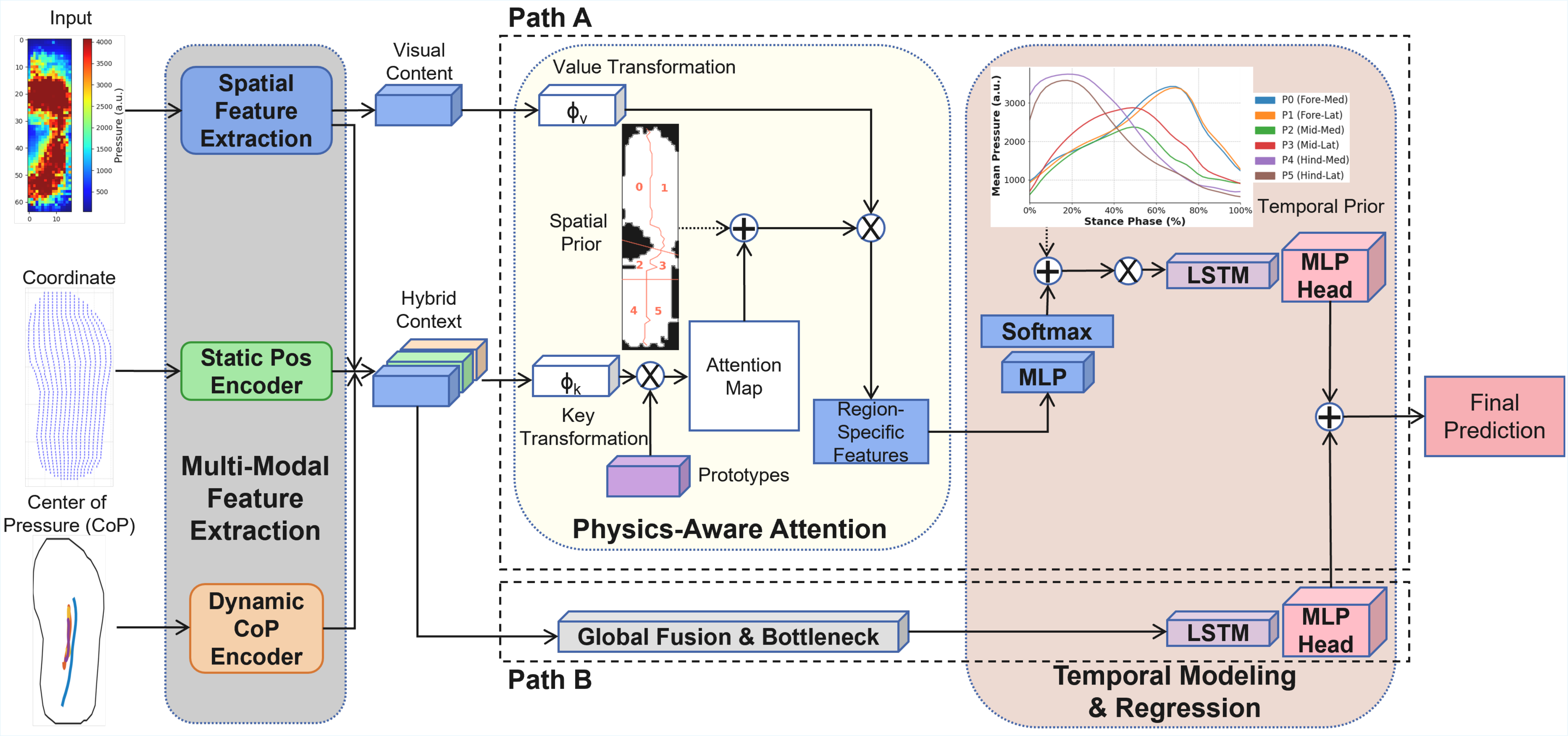}
    \caption{\textbf{The proposed Dual-Path Region-Guided Attention Network (DP-RGNet) architecture.} The model employs a shared spatial encoder feeding into two complementary pathways: the Attention Context Path (Path A) and the Global Feature Path (Path B).}
    \label{fig:dp-rgnet}
\end{figure}

\section{Proposed Model}
We propose the Dual-Path Region-Guided Attention Network (DP-RGNet), a novel architecture designed to leverage the high-resolution and spatio-temporal structure of plantar data measurements. DP-RGNet comprises a shared spatial encoder followed by two parallel and complementary pathways: a region-guided path (Path A) that organizes features according to anatomy-informed plantar regions and a global feature path (Path B) that processes globally aggregated representations without explicit regional constraints. The outputs of the two paths are combined to produce the final predictions. In the following, we use Dataset A as a representative example. The overall architecture is illustrated in \textbf{Figure 1}. The model takes the sequential pressure maps as input and outputs the corresponding time series signals of GRF/GRM. 

\subsection{Shared Spatial Encoder}
The initial processing is performed by a shared spatial encoder, which acts as the primary feature extractor for both paths. The encoder maps the raw pressure grid into a set of grid-wise feature vectors, providing a consistent representation of plantar loading patterns.

The input frames are first passed through a stack of 2D convolutions with batch norm layers, which captures local spatial structures in the pressure maps. The resulting high-level feature maps are then flattened along the spatial dimensions to obtain a sequence of features, forming the shared visual embedding used by both the region-guided and global paths.

\subsection{Dual-Path Architecture}
The shared features are subsequently processed through two parallel pathways to exploit complementary spatial and temporal representations.

\begin{figure}
    \centering
    \includegraphics[width=0.7\textwidth]{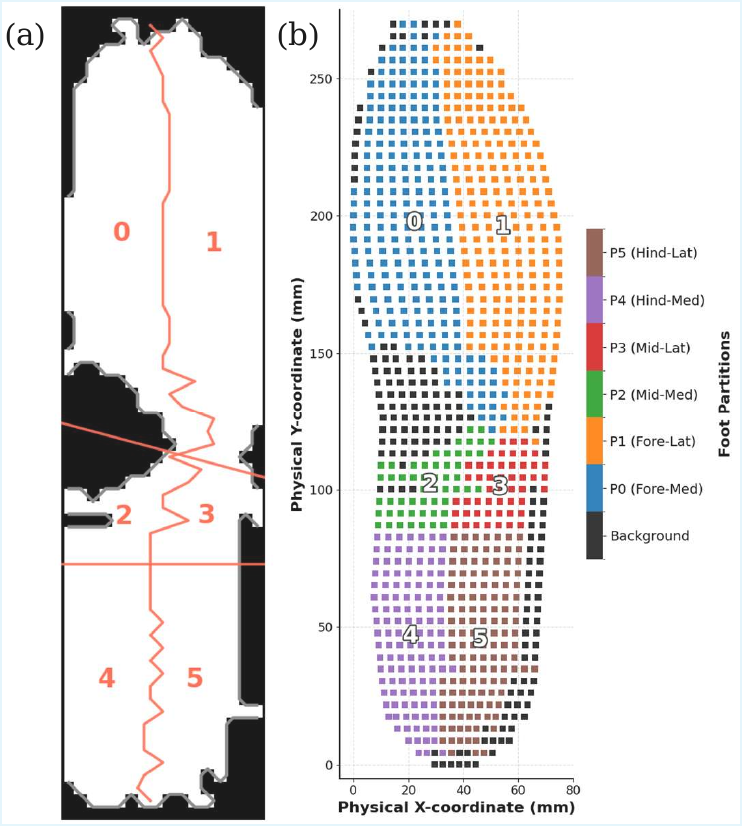}
    \caption{\textbf{Spatial Prior Definitions for Foot Segmentation} (a) Schematic diagram illustrating the six functional partitions (0-5) used for spatial segmentation. (b) The explicit mapping of the high-resolution sensor grid (1024 sensors) to their corresponding partition IDs, plotted by physical (mm) coordinates.}
\end{figure}

\begin{figure}
    \centering
    \includegraphics[width=0.75\textwidth]{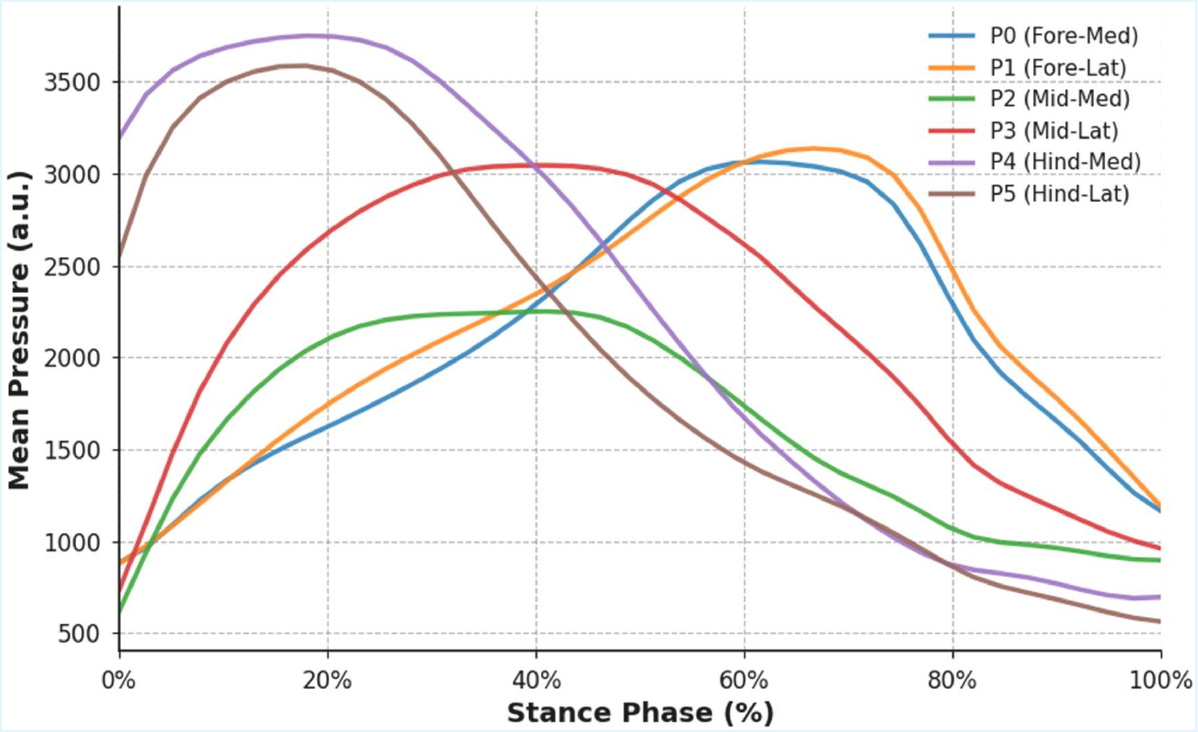}
    \caption{\textbf{Temporal Priors for Prototype Activation} The curves illustrate the mean pressure (in arbitrary units, a.u.) for each of the six functional prototypes, averaged across all subjects in the training dataset. The time axis is normalized during a complete stance phase.}
\end{figure}

\subsubsection{Region-Guided Context Path}
The first pathway corresponds to the region-guided branch of DP-RGNet and explicitly incorporates anatomical priors into the spatial aggregation of insole features. Since the sensor coordinates help the model to understand the non-Euclidean relationships of the sensor grids, and the Center of Pressure (CoP) has significant influence on GRF estimation\citep{shaulian2018effect}. We embedded features $Z_{\text{feat\_A}}$ with a combination of coordinate-based positional encodings and dynamic CoP encodings that vary over time, and then passed through a region-guided attention module. It integrates two types of priors:

\paragraph{1. Spatial Prior}
To enable the prototype-based pooling in the context path, we pre-define 6 anatomical regions of the foot as feature prototypes \citep{chen2021lisfranc}\citep{peicha2002anatomy}\citep{ribeiro2022effectiveness}. This partitioning is derived by analyzing the mean pressure distribution across the entire dataset. Initial Segmentation was accomplished by computing the mean pressure map across all collected samples to establish a general foot profile. We applied Otsu-based thresholding to isolate the regions of active foot contact. Then, the resulting contact area is divided into three major anatomical segments based on percentage length, $54\%$ for forefoot-midfoot boundary and $29\%$ from the bottom for midfoot-hindfoot boundary. From the other view, a row-specific medial-lateral midline is calculated based on the center of mass of active pressure in each row. The intersection of the anterior-posterior and medial-lateral boundaries yields the final six regions: the forefoot (lateral/medial), midfoot (lateral/medial), and hindfoot (lateral/medial). This anatomically constrained partition map is then used to provide a coarse anatomical segmentation of the foot and acts as a soft structural prior for region-guided attention, ensuring that the context path aggregates features according to established biomechanical principles. This regularizes the learning of the model, encorages region-level representations. \citep{PCMFootSegmentation} \textbf{Figure 2} shows the spatial definitions for 6 anatomical regions.
\paragraph{2. Temporal Prior}
The GRF/GRM signal is highly non-uniform over time, peaking sharply at specific phases of the gait cycle. This temporal non-uniformity serves as a strong temporal prior, which can be leveraged for advanced model training or analysis. We establish a Soft Regularization Attention Pattern that models the expected temporal sequence of contact intensity for the 6 anatomical prototypes across the normalized stance phase. With the predefined spatial segmentation map, we compute the characteristic activation curve for each prototype. The empirical activation $A_{t,k}$ for prototype $k$ at time step $t$ is calculated as the mean pressure of all active sensors within that region:$$A_{t,k} = \frac{1}{|\Omega_k|} \sum_{(h,w) \in \Omega_k} \bar{X}_{t,h,w}$$where $\Omega_k = \{(h,w) \mid \mathcal{M}_{h,w} = k\}$ is the set of spatial coordinates belonging to prototype $k$. This matrix is normalized to form the target attention tensor $\mathbf{P}$:$$\mathbf{P}_{t,k} = \frac{A_{t,k} + \epsilon}{\sum_{j=1}^K (A_{t,j} + \epsilon)}$$where $\epsilon$ is a small constant for numerical stability.  \textbf{Figure 3} illustrates the sequential nature of these empirical activation curves. $\mathbf{P}$ serves as a soft regularization target and explicitly guides the mechanism to attend to anatomically relevant prototypes during specific gait phases.

\paragraph{Region-Guided Attention} To obtain a compact and anatomically interpretable representation, DP-RGNet employs a region-guided attention mechanism that aggregates the local spatial features into a small set of region-level descriptors. At each time step, the pressure frame is first processed by the shared CNN encoder, producing grid-level pressure features. In parallel, the fixed sensor coordinates are embedded through a positional encoding module \citep{vaswani2017attention}, and the center of pressure trajectory is mapped to a dynamic positional encoding. Unlike standard static positional embeddings to represent fixed pixel coordinates, plantar pressure distribution exhibits high temporal variability centered around the CoP. To capture this dynamic spatial context, we introduce a Dynamic CoP Encoding module.
At each time step $t$, the instantaneous CoP coordinates $c_t = (x_{cop}, y_{cop})$ are mapped into a high-dimensional feature space using Fourier feature mapping $\gamma(\cdot)$:
\begin{equation}
\gamma(c_t) = [ \sin(2^0 \pi c_t), \cos(2^0 \pi c_t), \dots, \sin(2^{L-1} \pi c_t), \cos(2^{L-1} \pi c_t) ]
\end{equation}
The resulting Fourier features are then projected by a learnable Multi-Layer Perceptron (MLP) to obtain the final dynamic embedding $Z_{cop} \in \mathbb{R}^{d}$, which matches the dimension of the visual features. By explicitly injecting the CoP trajectory, the model learns to calibrate its spatial attention relative to the center of force at the moment.
These three components are concatenated along the feature dimension to form a fused spatial presentation $Z_{\text{feat}}$, which encodes local pressure patterns, spatial location on the plantar surface, and CoP information. Instead of treating all spatial locations equally, we adopted the attention mechanism inspired by the transformer\citep{vaswani2017attention}. We use a set of learnable prototype queries $Q \in \mathbb{R}^{R \times d}$, where each query corresponds to a specific anatomical region guided by the prior; The context features are projected to keys $K \in \mathbb{R}^{N \times d}$ through a linear projection layer $\phi_k$ and the content features are projected to values $V \in \mathbb{R}^{N \times d}$ through a linear projection layer $\phi_v$.
The core difference lies in the integration of a static anatomical prior, denoted as the bias matrix $\mathrm{Bias} \in \mathbb{R}^{R \times N}$, derived from the partition map. We denote by $N$ the number of insole sensors and by $R$ the number of anatomical regions. The attention scores are computed as:
\begin{equation}
\text{Attention}(Q, K) = \text{Softmax}\left( \frac{Q K^T}{\sqrt{d_k}} + \lambda Bias \right)
\end{equation}
where $Bias_{ij}$ is set to a positive bias value if the $j$-th sensor belongs to the $i$-th anatomical region, and $0$ otherwise. $d_k$ is the per-head key dimension. $\lambda$ is a scalar controlling the strength of the prior. This guides the model to learn how to focus on prior anatomical regions, while retaining the flexibility to attend to long-range dependencies through the data-driven dot-product term.
The resulting representation $Z$ is a set of region-level features for each time step, aggregating the values $V$ based on the attention scores:
\begin{equation}
Z = \text{Attention}(Q, K) V
\end{equation}
which would be the input of the following temporal modeling block.

\paragraph{Temporal Modeling} The region-level features are then reshaped into temporal sequences  and fed into a Bidirectional LSTM in the region-guided path. This recurrent module captures the evolution of region interactions over the gait cycle and produces temporal context vectors $H_{\text{region}}$. A final linear projection maps $H_{\text{region}}$ to the GRF/M predictions $\hat{y}_A \in \mathbb{R}^{B \times T \times 6}$ for this path.

\subsubsection{Global Feature Path}
The second pathway implements the global branch of DP-RGNet and captures more holistic spatial patterns that are not explicitly constrained by the anatomical priors. The shared features $Z_{\text{feat}}$ are embedded with coordinate and CoP features same to Path A. And then aggregated through a bottleneck projection, which compresses the full insole grid into a low-dimensional global feature vector at each time step. These global representations are reshaped into temporal sequences, denoted as $X_{global}$, and passed through a separate Bidirectional LSTM. The output of this recurrent module is the global temporal context $H_{global}$, which is finally mapped to the predictions $\hat{y}_B \in \mathbb{R}^{B \times T \times 6}$.

Finally, the outputs of the region-guided context path and the global feature path are summed to obtain the final GRF/M estimates:
$$
\hat{y} = \hat{y}_A + \hat{y}_B.
$$

\section{Experiments}
\subsection{Experiment Details}
We trained all models with an NVIDIA GeForce RTX 4090 GPU. All training processes were performed using Python (version 3.10), PyTorch (version 2.6.0). Mean squared error (MSE) was used as the loss function. AdamW was used as the optimizer. Cosine Annealing Learning scheduler was applied, with the period set to the number of epochs and the minimum learning rate of $5 \times 10^{-6}$ for dataset A and $1 \times 10^{-6}$ for dataset B. All models were trained for up to 60 epochs with early stopping (patience of 10 epochs). The results were reported in correlation coefficient (r), root mean square error (RMSE) and normalized RMSE (NRMSE). The reported metrics are computed based on the best model's weights with the lowest loss on the validation data.

\subsection{Comparison models}
To assess the effect of the proposed DP-RGNet architecture and its components, we compare four model variants under identical training conditions: (1) CNN, (2) CNN+LSTM, (3) DP-RGNet (full model), and (4) DP-RGNet (Path B only). Path A serves as an auxiliary branch, aggregating sensor elements into small number of semantic regions. Therefore, we do not consider Path A alone as a meaningful baseline for the main regression task. All comparison models are implemented and trained with the same data preprocessing, loss function, optimizer, scheduler and early-stopping protocol described in Experiment Details.

For Dataset A, which consists of fewer subjects and trials, all models were trained exclusively under the same, lower-capacity configuration of model parameters. This choice reduces effective model capacity and mitigates overfitting risks that usually happen with a small training set. Details are listed in \textbf{Table 2}.

For Dataset B, we evaluated a low-resolution configuration is obtained by downsampling the original grid by a factor of two in each spatial dimension and pairing it with a higher-capacity model. This was done to reduce the burden of computational demands. Details are listed in \textbf{Table 3}.

\begin{table}[hbtp]
\centering
\caption{Key Hyperparameters for models on Dataset A. These parameters define all the tested models.}
\begin{tabular}{l p{2.5cm} p{2.5cm} p{3.0cm}} 
\toprule
\textbf{Hyperparameter} & \textbf{DP-RGNet} & \textbf{Baseline Models} \\
\midrule
CNN Feature Dim & 128 & 128\\
Positional Encoding Dim & 128 & / \\
Feature Embedding Dim & 256 & / \\
CoP Encoding Dim & 128 & / \\
Bottleneck Dim & 128 & / \\
Regional LSTM Hidden Size & 256 & / \\
Global LSTM Hidden Size & 256 & 256(CNN+LSTM) \\
\midrule
LSTM Layers & 2 & 2\\
Dropout Rate & 0.35 & 0.2 \\
\bottomrule
\end{tabular}
\end{table}

\begin{table}[hbtp]
\centering
\caption{Key Hyperparameters for Low-Resolution and Baseline Model Configurations on Dataset B. These parameters define all the tested models.}
\begin{tabular}{l p{2.5cm} p{2.5cm} p{3.0cm}} 
\toprule
\textbf{Hyperparameter} & \textbf{Low-Res} & \textbf{Baseline Models} \\
\midrule
CNN Feature Dim & 128 & 128\\
Positional Encoding Dim & 128 & / \\
Feature Embedding Dim & 256 & / \\
CoP Encoding Dim & 256 & / \\
Bottleneck Dim & 256 & / \\
Regional LSTM Hidden Size & 256 & / \\
Global LSTM Hidden Size & 256 & 256(CNN+LSTM) \\
\midrule
LSTM Layers & 2 & 2\\
Dropout Rate & 0.3 & 0.1 \\
\bottomrule
\end{tabular}
\end{table}

The baseline CNN model is designed to isolate the contribution of temporal modeling and region-guided attention. It uses a convolutional encoder with the same depth and channel configuration as in the proposed framework. Coordinate- and CoP-based positional encodings with the same embedding dimensions as in DP-RGNet are also input to the CNN features. The fused features are then flattened over the spatial dimension and compressed into a bottleneck vector for each frame. This bottleneck vector is directly mapped to GRF/M via a regression head.

The baseline CNN+LSTM model extends the CNN architecture by adding 2-layer Bidirectional LSTM.

The Path-B-only variant of DP-RGNet uses the same convolutional encoder and the same coordinate and CoP positional encodings as the full DP-RGNet. The Path A is removed, and the final fusion between Paths A and B is omitted, so that only Path B is used to generate the final prediction.

\subsection{Evaluation}
For Dataset A, model performance was evaluated using 5-fold cross-validation at the step level, ensuring that all models were evaluated under identical data partitions. For Dataset B, we conducted 5-fold cross-validation at the subject-wise level. Participants were split into 5 non-overlapping subject-based folds. This protocol allows us to evaluate generalization to unseen subjects while maintaining a consistent evaluation procedure across models.

\section{Results}
\subsection{Dataset A}
\textbf{Table 4} summarizes the prediction performance of all compared models on Dataset A, evaluated by the correlation coefficient (R), root mean squared error (RMSE), and normalized RMSE (NRMSE). Across all six GRF/M components, DP-RGNet consistently outperforms the CNN and CNN+LSTM baselines, as well as the Path-B-only variant. \textbf{Figure 4} shows an example step of the ground truth and predicted values.

\begin{figure}
    \centering
    \includegraphics[width=1\textwidth]{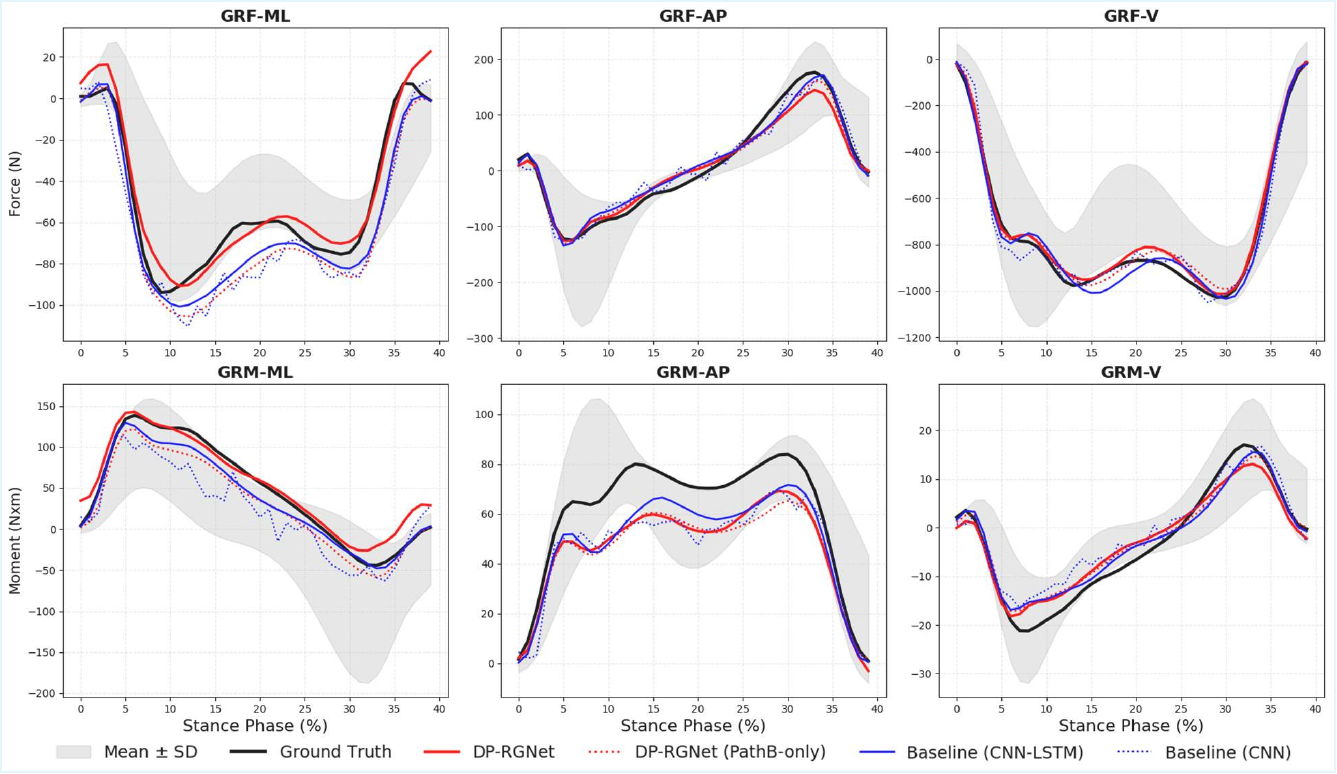}
    \caption{Measured (solid black line, with shaded gray area representing the mean $\pm$ SD) and predicted GRF/GRM components of the selected sample: DP-RGNet (red solid line), DP-RGNet (PathB-only) (red dot line), CNN-LSTM Baseline (blue solid line), and CNN Baseline (blue dot line) during the stance phase of an example step of subject 4 walking at 1.0 m/s.}
\end{figure}

Compared to the strongest baseline (CNN+LSTM), DP-RGNet achieves lower NRMSE and higher correlation on all force and moment components. The performance shows that incorporating anatomically informed region aggregation and dual-path temporal modeling improves the robustness and accuracy of GRF/M estimation in the small-sample insole setting.

\begin{table}[hbtp]
\small
\centering
\caption{Correlation coefficient ($r$), RMSE (mean (SD)), and NRMSE (mean (SD)) between the measured GRFs/GRMs (by force plate) and predicted GRFs/GRMs (by insole data) in Model CNN, CNN+LSTM, DP-RGNet (Path B-only) and DP-RGNet on Dataset A}
\medskip

\begin{tabular}{l ccc ccc}
\toprule
\textbf{Variable} & \multicolumn{3}{c}{\textbf{CNN}} & \multicolumn{3}{c}{\textbf{CNN+LSTM}} \\
& $r$ & RMSE ($N or Nm$) & NRMSE ($\%$) & $r$ & RMSE ($N or Nm$) & NRMSE ($\%$) \\
\midrule
\multicolumn{7}{l}{\textbf{Right Foot}} \\ 
\midrule
GRF\_ML & 0.945 (0.001) & 10.377 (0.156) & 10.84 (0.24) & 0.969 (0.001) & 8.107 (0.157) & 8.65 (0.09) \\
GRF\_AP & 0.976 (0.001) & 20.880 (0.426) & 6.20 (0.13) & 0.989 (0.001) & 13.914 (0.287) & 4.29 (0.06) \\
GRF\_V & 0.970 (0.001) & 73.443 (0.931) & 7.07 (0.07) & 0.986 (0.001) & 47.558 (0.825) & 4.65 (0.06) \\
GRM\_ML & 0.971 (0.002) & 19.548 (0.675) & 9.39 (0.30) & 0.988 (0.002) & 14.170 (0.535) & 6.92 (0.27) \\
GRM\_AP & 0.962 (0.002) & 8.792 (0.135) & 10.05 (0.34) & 0.981 (0.002) & 6.851 (0.157) & 8.04 (0.40) \\
GRM\_V & 0.969 (0.002) & 2.900 (0.060) & 7.49 (0.14) & 0.985 (0.001) & 2.090 (0.048) & 5.57 (0.11) \\
\bottomrule
\end{tabular}

\medskip 

\begin{tabular}{l ccc ccc}
\toprule
\textbf{Variable} & \multicolumn{3}{c}{\textbf{DP-RGNet (Path B-only)}} & \multicolumn{3}{c}{\textbf{DP-RGNet}} \\
& $r$ & RMSE ($N or Nm$) & NRMSE ($\%$) & $r$ & RMSE ($N or Nm$) & NRMSE ($\%$) \\
\midrule
\multicolumn{7}{l}{\textbf{Right Foot}} \\ 
\midrule
GRF\_ML & 0.970 (0.007) & 7.847 (0.092) & 8.33 (0.09) & 0.972 (0.002) & 7.550 (0.126) & 8.06 (0.19) \\
GRF\_AP & 0.990 (0.001) & 13.334 (0.207) & 4.12 (0.09) & 0.991 (0.001) & 12.359 (0.311) & 3.85 (0.08) \\
GRF\_V & 0.986 (0.001) & 46.003 (0.663) & 4.50 (0.09) & 0.987 (0.001) & 42.453 (0.731) & 4.18 (0.06) \\
GRM\_ML & 0.989 (0.001) & 13.264 (0.442) & 6.51 (0.21) & 0.990 (0.001) & 12.198 (0.502) & 5.99 (0.26) \\
GRM\_AP & 0.982 (0.002) & 6.615 (0.123) & 7.73 (0.40) & 0.983 (0.002) & 6.403 (0.085) & 7.52 (0.37) \\
GRM\_V & 0.985 (0.001) & 2.018 (0.016) & 5.37 (0.08) & 0.987 (0.001) & 1.907 (0.049) & 5.10 (0.12) \\
\bottomrule
\end{tabular}
\end{table}

\begin{table}[hbtp]
\small
\centering
\caption{Correlation coefficient ($r$), RMSE (mean (SD)), and NRMSE (mean (SD)) between the measured vertical GRF and predicted vertical GRF (by pressure sensing walkway) in Model CNN, CNN+LSTM, DP-RGNet (Path B-only) and DP-RGNet on Dataset B}
\medskip

\begin{tabular}{l ccc ccc}
\toprule
\textbf{Variable} & \multicolumn{3}{c}{\textbf{CNN}} & \multicolumn{3}{c}{\textbf{CNN+LSTM}} \\
& $r$ & RMSE ($N$) & NRMSE ($\%$) & $r$ & RMSE ($N$) & NRMSE ($\%$) \\
\midrule
\multicolumn{7}{l}{\textbf{Right Foot}} \\ 
\midrule
GRF\_V & 0.997 (0.001) & 14.052 (3.447) & 2.59 (0.74) & 0.998 (0.001) & 9.262 (1.246) & 1.60 (0.23) \\
\bottomrule
\end{tabular}

\medskip

\begin{tabular}{l ccc ccc}
\toprule
\textbf{Variable} & \multicolumn{3}{c}{\textbf{DP-RGNet (Path B-only)}} & \multicolumn{3}{c}{\textbf{DP-RGNet}} \\
& $r$ & RMSE ($N$) & NRMSE ($\%$) & $r$ & RMSE ($N$) & NRMSE ($\%$) \\
\midrule
\multicolumn{7}{l}{\textbf{Right Foot}} \\ 
\midrule
GRF\_V & 0.998 (0.001) & 11.144 (2.864) & 1.93 (0.50) & 0.999 (0.001) & 8.498 (1.060) & 1.42 (0.13) \\
\bottomrule
\end{tabular}

\end{table}

\subsection{Dataset B}
On Dataset B, we evaluated the same set of models under the subject-wise 5-fold cross-validation. \textbf{Table 5} reports the results for the vertical GRF component, which is the only ground-truth channel available in this public dataset. All models obtain relatively low NRMSE on this task, reflecting the more regular gait patterns and larger number of steps in Dataset B. However, DP-RGNet still surpasses the CNN and CNN+LSTM baselines, demonstrating that the proposed architecture does not sacrifice performance when sufficient data are available and the prediction target is less challenging.

\section{Discussion}

This study investigated the potential of using anatomically guided attention to improve the estimation of ground reaction forces and moments from plantar pressure measurements, especially when only a limited number of subjects is available. The proposed Dual-Path Region-Guided Attention Network (DP-RGNet) consistently outperformed conventional CNN and CNN+LSTM baselines, as well as a Path-B-only variant, on our insole dataset (dataset A) with four subjects and 2,343 steps. Despite the small number of participants, DP-RGNet achieved notably lower NRMSE than other compared models. The performance improvements over the CNN+LSTM and Path-B-only variants indicate that the dual-path architecture, and in particular the region-guided branch, contributes information that cannot be recovered by global or temporal modeling alone. These results suggest that explicitly encoding anatomical structure and region-level dynamics provides a useful inductive bias for small-sample gait analysis\citep{karatsidis2016estimation}\citep{dziuk2023peak}\citep{revi2020indirect}.

Importantly, the priors used in this work are not exact physical models. The spatial region priors and the temporal attention priors are distributed from techniques based on the average plantar pressure, rather than from subject-specific calibration or detailed biomechanical simulations. As a consequence, these priors inevitably contain mismatches with a given individual step. However, empirical results show that such approximate priors are still highly beneficial. The prior masks in DP-RGNet are used as soft biases in the attention mechanism rather than as hard constraints. They encourage the network to align its region-level aggregation with anatomically plausible partitions and with typical loading patterns, while still allowing the learned attention weights to deviate when supported by the data. This design strikes a balance between incorporating domain knowledge and preserving flexibility for data-driven adaptation.

In addition, we tested all models on a public force-plate dataset (dataset B) that provides only vertical GRF signals. On dataset B, all methods achieved relatively low NRMSE, with DP-RGNet slightly exceeded the baselines. Although the improvement is smaller compared to the insole dataset, these results show that the proposed architecture maintains superior performance in a laboratory scenario where only vertical GRF data is measured. This suggests that the auxiliary positional encoding and prior anatomical segmentation provide more robust features\citep{matsumoto2024comparing}\citep{joo2016improving} to gait analysis by offering structural constraints compared to traditional baseline models.

In clinical diagnosis, the differentiation between healthy and pathological gait is often based on subtle deviations. For example, previous studies examining patients with unilateral anterior cruciate ligament reconstruction (ACLR) found that the asymmetry index in vertical GRF during walking is significantly different between healthy subjects and patients, particularly at the second peak force ($p \le 0.001$) \citep{mantashloo2020vertical}. The natural asymmetry in healthy subjects is generally below $6\%$ for vGRF. A baseline model with a higher error margin exceeding this threshold makes it less reliable to distinguish true pathological changes from noise. In contrast, reducing the error to the performance of the proposed model provides a safer margin, which is critical for improving the clinical fidelity required for reliable detection of the subtle biomechanical compensatory strategies required following ACLR.

Many recent works have focused on IMU-based GRF estimation and other applications for gait analysis \citep{shahabpoor2018estimation}\citep{caramia2018imu}\citep{liu2023imu}. IMUs are inexpensive and widely available, but provide only indirect information about plantar loading patterns. Instrumented insoles, by contrast, directly measure plantar pressure and are well suited for long-term, in-shoe deployment in daily life. Our results demonstrate that, when combined with an appropriate deep learning architecture such as DP-RGNet, high-density insole measurements can approximate force-plate-level information for all six GRF/M components. This provides a complementary pathway to IMU-based methods and supports the broader goal of moving laboratory-grade gait analysis into real-world environments. This work demonstrates the feasibility of convenient, unsupervised gait monitoring in future applications. It could be further developed in personalized rehabilitation, for example, calculating Required Coefficient of Friction (RCOF) from estimated GRF to quantify and evaluate fall risk \citep{yamaguchi2017required}.
    
However, this study has several limitations. The insole dataset comprises only four subjects, although each contributed multiple walking speeds and a substantial number of steps. While this setting is representative of many practical data collection scenarios, it limits the diversity of anthropometrics and gait patterns, and therefore the generalizability of the quantitative results. Second, the public dataset used for external evaluation contains only vertical GRF, preventing a comprehensive comparison for shear forces and moments. Although both dataset measure the plantar pressure distribution, the data distribution could have a difference on these two datasets. As noted above, the spatial and temporal priors in DP-RGNet are currently derived from group-average statistics. More systematic analyses of prior design, including subject-adaptive priors, context-dependent priors (e.g., speed, the shape of foot), could further clarify the role of prior information and potentially yield additional gains. Also, implementing a simple learnable linear module, where the final prediction $\hat{y}$ is a weighted sum ($\hat{y} = w_A \hat{y}_A + w_B \hat{y}_B$), would explicitly quantify the relative contribution of the anatomical prior (Path A) versus the global features (Path B). Finally, although the present architecture is already structured to reduce the number of parameters in small-sample settings, further work is needed on model compression and on-device deployment. Lightweight variants of DP-RGNet and more efficient inference on embedded hardware would be an important step toward easily accessible insole-based gait analysis in daily life \citep{shahabpoor2018estimation}\citep{song2024estimating}.

\bibliographystyle{unsrtnat}


\end{document}